# Multi-Task Multi-Fidelity Learning of Properties for Energetic Materials

Robert J. Appleton[1], Daniel Klinger[2], Brian H. Lee[1], Michael Taylor[3], Sohee Kim[4], Samuel Blankenship[1], Brian C. Barnes[5], Steven F. Son[2], Alejandro Strachan[1*]

[1] School of Materials Engineering and Birck Nanotechnology Center, Purdue University, West Lafayette, Indiana 47907, USA

[2] School of Mechanical Engineering, Purdue University, West Lafayette, Indiana 47907, USA

[3] Department of Mechanical Engineering, New Mexico Institute of Mining and Technology, Socorro, New Mexico 87801, USA

[4] Department of Computer Sciences, Purdue University, West Lafayette, Indiana 47907, USA

[5] U.S. Army Combat Capabilities Development Command Army Research Laboratory, Aberdeen Proving Ground, Maryland 21005, USA

## Abstract

Data science and artificial intelligence are playing an increasingly important role in the physical sciences. Unfortunately, in the field of energetic materials data scarcity limits the accuracy and even applicability of ML tools. To address data limitations, we compiled multi-modal data: both experimental and computational results for several properties. We find that multi-task neural networks can learn from multi-modal data and outperform single-task models trained for specific properties. As expected, the improvement is more significant for data-scarce properties. These models are trained using descriptors built from simple molecular information and can be readily applied for large-scale materials screening to explore multiple properties simultaneously. This approach is widely applicable to fields outside energetic materials.

* Corresponding author: strachan@purdue.edu

## 1. Introduction

The selection of energetic materials (EM) for specific applications and the discovery and exploration of new materials is a multi-objective problem[1–5]. Key among several material characteristics to consider are detonation properties and sensitivity metrics[1,3,4]. Unfortunately, obtaining these quantities experimentally is costly in terms of both resources and time. Thus, easy-to-evaluate and accurate models can significantly accelerate the discovery of new materials and down-select candidates to be tested experimentally. Historically, analytical expressions for detonation properties were derived from thermochemical models and physicochemical domain expersite[6–9]. More recently, physics-based modeling across scales have contributed to our understanding of the shock and thermal initiation of chemistry in energetic materials[10–17], the shock to deflagration transition[10,17,18], their performance[11,19–21], the effect of microstructure on localization of energy and detonation initiation under shock loading[11,22–30]. Despite this progress, prediction of the shock to detonation transition or sensitivity (e.g. drop tests) requires significant parameterization against experiments[22,23,31–35]. In addition, these models are computationally intensive and restricted to expert users. Machine learning models could fill this gap in our predictive capabilities. Using ML, a combination of high-throughput computational data and deep learning achieved high accuracy in the prediction of detonation velocity and detonation pressure[36,37]. Unfortunately, properties like sensitivity are known for a much smaller number of materials[38,39]. For such cases, techniques such as transfer learning[40] and feature engineering[38,41–44] have shown promise.

The challenge of data scarcity is not unique to the field of energetics, it is pervasive in materials science[45–47]. In most cases, we have large datasets of low-fidelity data[48] while the datasets for high-fidelity data[49,50] are scarce. To address this problem, recent research has been focused on the design of models capable of leveraging the abundance of low-fidelity data to help improve predictability on scarce high-fidelity data[45–47,51]. In the context of our dataset, we refer to high-fidelity as the experimental data and low-fidelity as the calculated data. In this paper we address another important challenge, the fact that while some properties are known or can be computed for most materials, other critical data is only known for a relatively small subset. To maximize the information available to predictive models for all these properties, we use multi-task learning. Here, we train a single model that learns about all the materials available and their properties. A key approach tested in this paper is to add a *selector* input that determines the output of interest and train a single model. In this way, the model can learn common abstractions for each material from the entire dataset and use them to predict the various outputs. This technique has been successfully used for polymer by Kuenneth et al.[45,46] who demonstrated the ability to learn up to 36 different properties simultaneously using data that came from sparse datasets. We apply these multi-task multi-fidelity neural networks to properties of interest for energetic materials and evaluate the performance of each model by comparing to benchmark single-task models developed using random forests and traditional dense neural networks. We find that the multi-task multi-fidelity neural networks outperform the single-task benchmark models.

## 2. Data

We focus on the following properties of energetic materials: detonation velocity ($D$), detonation pressure ($P$), heat of detonation ($Q_{ex}$), Gurney energy ($E_G$), impact sensitivity (drop height ($h_{50}$) and drop energy ($E_{50}$)), crystal heat of formation ($H_f$), heat of sublimation ($H_{sub}$), and gas heat of formation ($H_{gas}$). We collected experimental data for detonation velocity, detonation pressure, heat of detonation, impact sensitivity (drop height and drop energy), and crystal heat of formation from the following refs[38,39,52–54]. In addition, we use theoretical data for detonation velocity, detonation pressure, heat of detonation, Gurney energy, impact sensitivity (drop-height), heat of sublimation, and gas heat of formation published in Refs. [39,54,55]. The total amount of data of each property and the method for the data collection are summarized in Table 1.

*Table 1. Property datasets used in this study. **a)** experimental sources and **b)** theoretical sources. The total number of datapoints for each property is listed along with the percentage of unique materials that have that property.*

| a) Experimental Properties | Datapoints | Method |
|---|---|---|
| detonation velocity ($D$)[52,54] | 262 (25.5%) | Various experimental testing methods |
| detonation pressure ($P$)[52,54] | 66 (6.4%) | Various experimental testing methods |
| heat of detonation ($Q_{ex}$)[52] | 68 (6.6%) | Various experimental testing methods |
| impact sensitivity drop height ($h_{50}$)[39] | 306 (29.8%) | Drop-weight impact test |
| crystal heat of formation ($H_F$)[52] | 236 (23.0%) | Originally sourced from ICT[56] and NIST[57] |
| b) Calculated Properties | Datapoints | Method |
| detonation velocity ($D$)[39,54,55] | 491 (47.8%) | Empirical equations / Thermochemical codes |
| detonation pressure ($P$)[39,54,55] | 444 (43.2%) | Empirical equations / Thermochemical codes |
| heat of detonation ($Q_{ex}$)[39] | 306 (29.8%) | DFT - Molecular fragment model |
| Gurney energy ($E_G$)[39] | 306 (29.8%) | $C_v$ model[58] |
| heat of sublimation ($H_{sub}$)[39] | 306 (29.8%) | Molecular fragment model |
| gas heat of formation ($H_{gas}$)[39] | 306 (29.8%) | DFT |

In total, we collected 3,097 datapoints across 643 unique materials, with a large fraction of the materials only having one or two properties. We only considered pure CHNOClF molecules and thus our dataset does not contain any composite mixtures or metals. To evaluate the correlations between the properties in our dataset we computed the Pearson correlation between each pair of properties using the datapoints that overlap. Figure 1 shows the Pearson coefficient (color) and the number in the square is the number of materials that are shared between the datasets. We note the strong positive correlation between experimental and theoretical detonation properties and the strong negative correlation between the detonation properties and the impact sensitivity data. The correlations between the enthalpic properties and the detonation properties or the sensitivity data are not as strong however we still want to leverage the addition of this data. A model capable of learning from all this data cumulatively is going to benefit from these underlying correlations between the properties of interest.

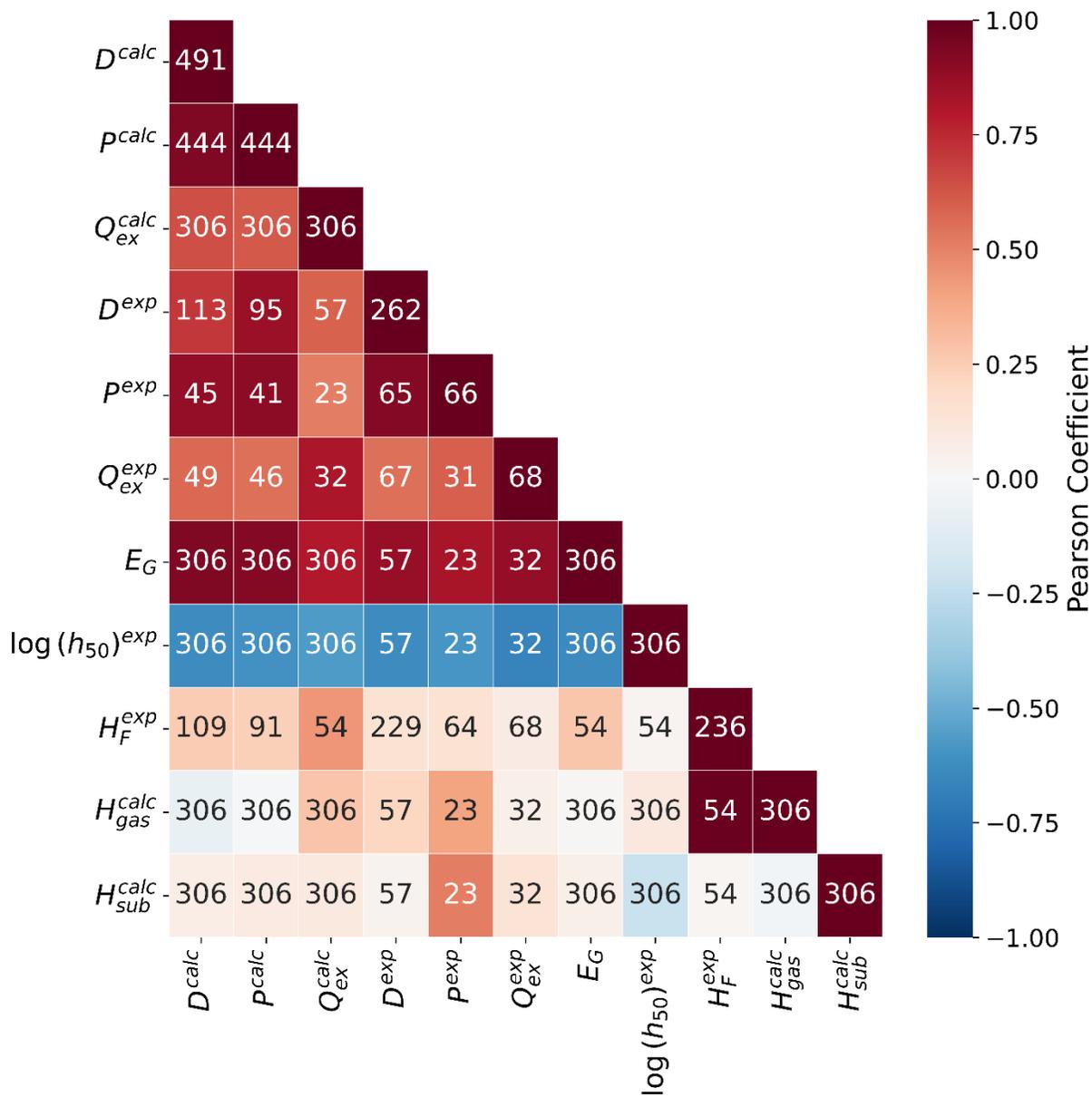

*Figure 1:* Pearson correlations between the different properties in the combined energetic materials dataset. The number in each square represents the number of materials that have both properties.

## 3. Machine learning models

**Modeling Objectives.** The goal of this work is to evaluate multi-task learning as a strategy to deal with multi-modal data and scarcity. We hypothesize that this approach will extract common materials features from the entirety of the data and use them to create accurate models for the various properties. Important choices include the representation of each material (featurization) and model selection (architecture). For each property dataset, we will create single-task models as benchmarks, serving to quantify the changes in predictive capabilities when employing multi-task learning models, as depicted in Figure 2(b-d). The subsections below discuss featurization and model architecture.

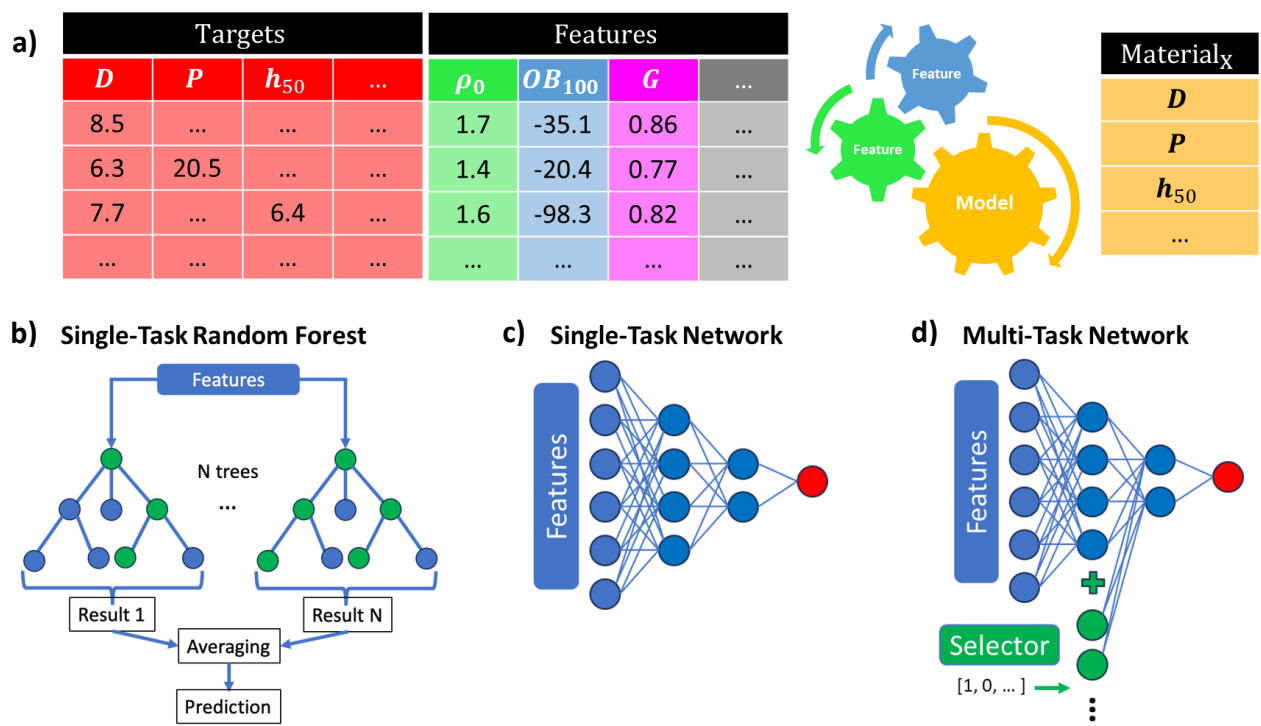

*Figure 2: Model designs for property prediction. **a)** Ideal scenario of taking sparse dataset of targets with features for each material and predicting all properties of interest. **b)** Single-task random forest for property prediction benchmark. **c)** Single-task neural network for property prediction benchmark. **d)** Multi-task neural network that ingests all the data and uses the selector vector to inform the model which property it is predicting.*

**Featurization scheme**. The predictability of a machine learning model is inherently connected to how well the features capture the necessary information needed to predict the property of interest[59]. We use SMILES strings to represent the molecular structure of each material as a graph. From this graph, we can extract features using a combination of tools from the *mmltoolkit* python package[42], RDKit[60], and Mordred[61]. Inspired by previous benchmarks for ML modeling of EM properties [41–43,59], we implemented the following descriptors:

1. Electrotopological State Vector (ESV)[62]: The ESV descriptor is calculated by considering the topology of a molecule, which includes information such as the number of atoms, bonds, and cycles in the molecule, as well as their connectivity. The electrostatic properties of the molecule are also considered, including partial charges and atomic electronegativity values. The ESV descriptor is represented as a vector of numerical values, with each element corresponding to a specific topological or electrostatic property.
2. Functional group counts: From each molecule we count the number of functional groups (i.e., amino, azide, nitro groups).
3. Sum over bonds descriptor: First, all the molecules in the dataset are analyzed and all the distinct bond types are recorded. Then, for each molecule the number of each bond type present in that molecule is represented as an element in a vector.
4. Atomic counts and ratios: The ratio between nitrogen and carbon ($n_N/n_C$). Also, counts of the number of hydrogen and fluorine atoms are included.
5. Oxygen balance: Oxygen reacting with carbon and hydrogen is the main contributor to the energy release during explosion and there is a known relationship between the oxygen balance and the detonation velocity[63]. For that reason, we also use the oxygen balance defined as:

$$OB_{100} \equiv \frac{100}{n_{atoms}}\left(n_O - 2n_C - \frac{n_H}{2}\right)$$

where $n_C$, $n_H$, and $n_O$, are the number of carbon, hydrogen, and oxygen atoms in the molecule and $n_{atoms}$ is the total number of atoms in the molecule (this definition excludes consideration of metals).

To go beyond this, we added the following descriptors based on their representation of molecular structure, physiochemical relevance, and physical intuition:

1. Ring counts: Counts of rings of different sizes, as well as counts that differentiate between the type of ring (i.e., hetero, aliphatic, and aromatic).
2. Atomic and bond features: Number of rotatable bonds and aromatic atoms/bonds in each molecule. Number of hydrogen bond acceptors/donors. Atomic/bond polarization.
3. Acidic/basic group counts.
4. Atomic and Bond Contributions (ABC) of van der Waals volume of the molecule[64]: widely used descriptor in modeling physiochemical properties.
5. Weight ratio between the explosive gas products and the molecular weight of the explosive (G) as predicted from the $H_2O - CO_2$ decomposition assumption[6–9]: Using the $H_2O - CO_2$ decomposition assumption:

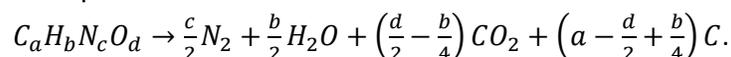
$$C_aH_bN_cO_d \rightarrow \frac{c}{2}N_2 + \frac{b}{2}H_2O + \left(\frac{d}{2} - \frac{b}{4}\right)CO_2 + \left(a - \frac{d}{2} + \frac{b}{4}\right)C.$$

We can estimate this ratio by the following equation:

$$G = \frac{56c + 88d - 8b}{48a + 4b + 56c + 64d}.$$

The datasets for each property include a corresponding density measured experimentally or predicted from a physics-based model. We train and evaluate models that use this density as an added descriptor as well as models that only use the molecular descriptors.

**Single-task models.** For each output, we fit two single-task (ST) models, a random forest (ST-RF) and a neural network (ST-NN). The performance of these models will be used as a benchmark to assess the multi-task neural networks (MT-NN). Schematics for the random forest and the neural network are shown in Figure 2(b,c). The RFs are built using the *scikit-learn*[65] package and the NNs are built using the *keras*[66] and *tensorflow*[67] packages.

**Multi-Task Models**. Traditional multi-task (MT) models have architectures with multiple outputs and are trained with datasets where each entry has all the outputs. However, it is often the case that not every output is available for every entry. To overcome this challenge, we implemented a custom multi-task neural network (MT-NN) adapted from Refs. [45,46]. The model has a single output and adds a selector vector that indicates what output property to predict to the set of input descriptors. The selector vector is a one-hot encoding of the possible outputs. A schematic of the general framework for the MT-NN is shown in Figure 2(d). The selector vector is concatenated to one of the hidden layers of the network, this location is a hyperparameter that is optimized for each MT-NN model. In this work, we found that most of the optimized MT-NN's were found to have the selector vector concatenated at the last hidden layer of the network. Only two out of fourteen models optimized to have the selector vector concatenated at the second to last hidden layer (see Table S3 in the Supplemental Material). A more detailed description of these models is presented in the Refs. [45,46]. As mentioned above, the idea of this approach is that the model will develop general descriptors of the problem at hand from the entirety of the data, and then use these descriptors to predict the various outputs (as indicated by the selector).

We apply the MT-NN approach to model various properties of interest for energetic materials from combined datasets that include disparate properties and theoretical and experimental data. We tested the MT-NN performance on seven subsets of output properties:

1. Detonation Properties Only: $D$, $P$, $Q_{ex}$ (calc and exp) and $E_G$ (calc).
2. Detonation and Sensitivity Properties: $D$, $P$, $Q_{ex}$ (calc and exp), $E_G$ (calc), and $h_{50}$ (exp).
3. Detonation and Thermodynamic Properties: $D$, $P$, $Q_{ex}$ (calc and exp), $E_G$ (calc), $H_{sub}$ (calc), $H_{gas}$ (calc), and $H_F$ (exp).
4. Thermodynamic Properties Only: $H_{sub}$ (calc), $H_{gas}$ (calc), and $H_F$ (exp).
5. Sensitivity and Thermodynamic Properties: $h_{50}$ (exp), $H_{sub}$ (calc), $H_{gas}$ (calc), and $H_F$ (exp).
6. All Properties

**Training**. For each model, we implement a 5-fold cross validation scheme to optimize the hyperparameters for each model. Then using the optimized parameters, we train and evaluate the model using an independent 5-fold cross validation scheme repeated 3 times with a different random seed for the data splitting. The hyperparameters of the ST models were optimized using *GridSearchCV* from the *scikit-learn*[65] package. The hyperparameters of the MT-NNs were optimized using *HyperBand*[68] from *keras*[66]. The resulting hyperparameters for each model are listed in Tables S1-S3 of the Supplemental Material.

## 4. Model Performance

Figure 3 compares the performance of the various single-task and multi-task models on the cross-validation test set for various properties; these models use the density as an input. Each bar represents the root mean squared error ($\overline{RMSE}$) averaged across the test sets from cross-validation and the three different random seeds (15 total points). We intentionally colored the ST-RF grey and separated it from the other models to emphasize the improvement of accuracy specifically between the ST-NN and MT-NN for each property.

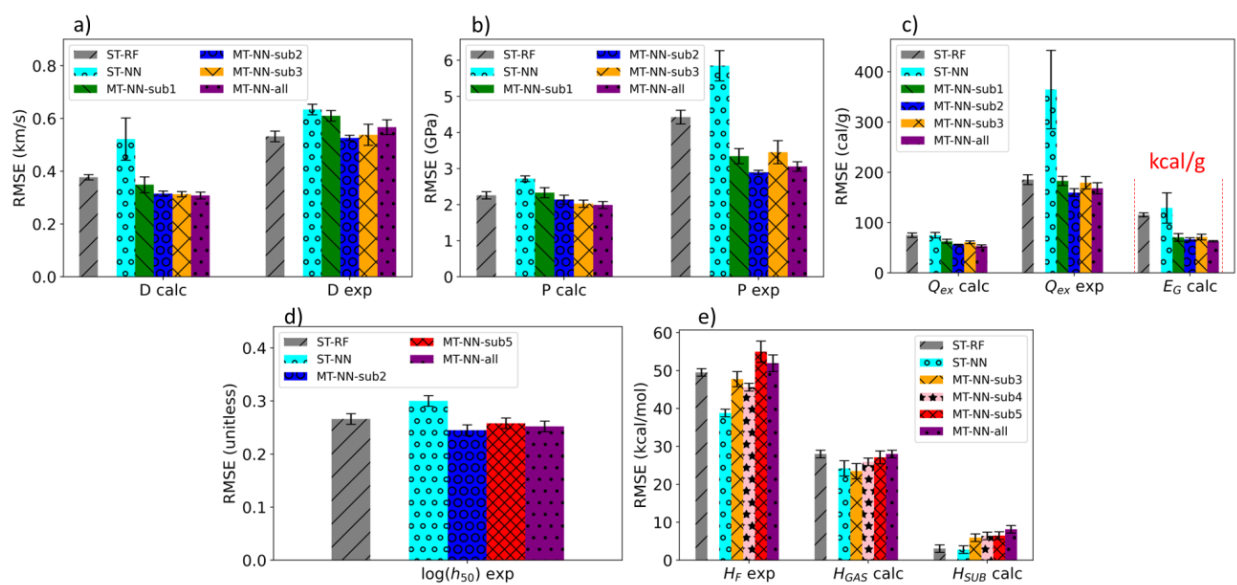

*Figure 3: Comparison of $\overline{RMSE}$ on the testing data for each model averaged across the 5-fold cross-validation and the three random seeds. The error bars represent the standard deviation for the test $RMSE$.*

The MT-NN models outperform both the ST-RF and the ST-NN models in terms of test $\overline{RMSE}$ for most properties, except the calculated heat of sublimation ($H_{sub}$ calc) and experimental crystal heat of formation ($H_F$ exp). We also note that the model trained on all 12 output properties (purple-colored bars with solid dot hatching) is the best for several properties. The properties that show the most significant improvement from multi-task learning are the experimental detonation pressure ($P$ exp) (30.2% reduction in $\overline{RMSE}$ compared to the best ST model) and experimental heat of detonation ($Q_{ex}$ exp) (13.9% reduction in $\overline{RMSE}$ compared to the best ST model), where the data is extremely limited (< 70 datapoints). The accuracy for the calculated heat of sublimation ($H_{sub}$ calc) and experimental crystal heat of formation ($H_F$ exp) is still comparable between the MT and ST models, and we attribute the lack of improvement from multi-task learning to the fact that these properties are not as strongly correlated with the rest (see Figure 1). We also note an improvement in the accuracy associated with safety (drop-weight sensitivity) which is important since these tests are not simple and require relatively large amounts of material. The results for experimental detonation properties also serve as a benchmark for ML prediction as this represents the largest database of experimental detonation properties used for model development to our knowledge, with approximately twice the number of experimental detonation velocities and triple the number of unique explosives as previously reported studies[44,69]. We see that our error in predicting experimental detonation velocity (test $\overline{RMSE}$ of 0.53 km/s for our best model) is higher than previous works[44,69] (test $RMSE$ of 0.23 - 0.28 km/s) and we attribute this to the fact our dataset includes a much higher diversity in the materials it contains, specifically materials containing Cl and F atoms. We also do not have multiple measurements from the same material where previous studies included different experimental detonation velocities at different densities.

Accurate estimates for the crystalline density of organic compounds are challenging to derive theoretically when their crystal structure is not known. Therefore, models based purely on descriptors derivable from the molecular skeleton are desirable. Since detonation properties depend strongly on density, see example Ref. [12], models lacking density as an input perform poorly. Interestingly, models trained without density perform well on sensitivity and thermodynamic properties, see Figure 4.

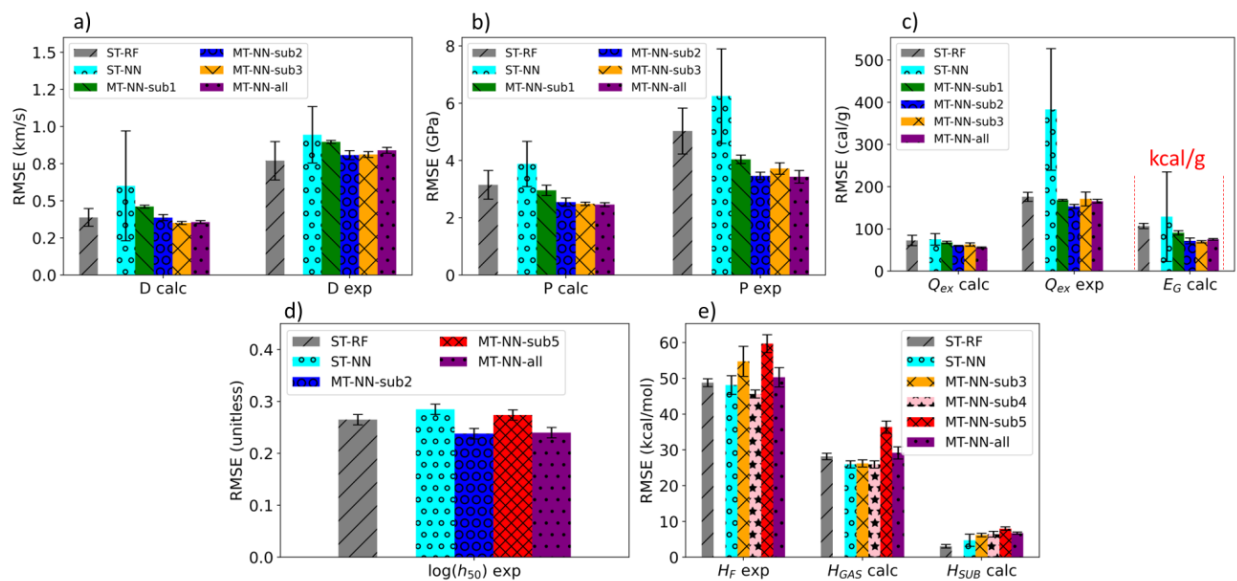

**Figure 4:** Comparison of $\overline{RMSE}$ on the testing data for each model averaged across the three random seeds. The error bars represent the standard deviation for the test RMSE.

As before, we find that the MT-NN models outperform the ST-RF and ST-NN models in terms of test $\overline{RMSE}$ for most properties, except for the calculated heat of sublimation ($H_{sub}$ calc). It is important to note, that the performance of these models on the experimental detonation properties, specifically $D$ and $P$, is significantly worse than the results that included density as a descriptor. We found that the best model for experimental drop-weight sensitivity ($log(h_{50})$ exp) is a MT-NN model that uses only the molecular descriptors and was trained with subset 2 (detonation and sensitivity properties: $D$, $P$, $Q_{ex}$ (calc and exp), $E_G$ (calc), and $h_{50}$ (exp)). This model can be very useful for high-level screening of new energetic materials because it allows for predictions of impact sensitivity from only the SMILES string and no other prior knowledge.

Our MT model outperforms the ST ones developed in this work and exhibits comparable accuracy to a recently published graph-based model[40] developed with the same dataset of impact sensitivities, see Table 2 below. The latter is based on *Chemprop*[70,71], a directed message passing neural network (D-MPNN) framework specifically designed for working on molecules. These graph models were shown to consistently outperform traditional feed-forward neural networks (FNN) for a wide variety of properties[70]. Therefore, we find it interesting that our FNN MT-NN can achieve comparable performance to the model presented in Ref. [40] by leveraging multi-task learning.

We note that the MT-NN trained on all 11 properties using only molecular descriptors performs only slightly worse. However, what we find more interesting is that the MT-NN trained with subset 2 which includes the molecular descriptors and the density as input features, was found to be noticeably less accurate. This indicates that though the density is a critical input feature for predicting the detonation properties of these materials it can hurt the predictability of the sensitivity.

**Table 2:** *Comparison of predictive accuracy on the experimental $log(h_{50})$ data between recently published graph-based model and models from current work.*

| Model | Test $\overline{RMSE}$ | Test $\overline{R^2}$ |
|---|---|---|
| Lansford Chemprop model[40]* | 0.233 ± 0.008 | 0.712 ± 0.013 |
| MT-NN-sub2 (molecular descriptors only) | 0.238 ± 0.010 | 0.705 ± 0.011 |
| MT-NN-all (molecular descriptors only) | 0.240 ± 0.010 | 0.700 ± 0.007 |
| MT-NN-sub2 (density + molecular descriptors) | 0.245 ± 0.010 | 0.689 ± 0.009 |
| ST-RF (molecular descriptors only) | 0.270 ± 0.010 | 0.615 ± 0.012 |
| ST-NN (molecular descriptors only) | 0.290 ± 0.010 | 0.598 ± 0.010 |

*Co-trained with 9,000 normalized electronic energies computed by the ANI-1cxx force field.

## 5. Conclusion

We applied custom multi-task neural networks for the prediction of several properties of interest for the design of energetic materials with different levels of fidelity. Our comparisons between different subsets of multi-task multi-fidelity models and single-task benchmark models show that multi-task learning enhances the predictive power for most properties. These results also demonstrate that multi-task learning is particularly helpful for properties where the data is limited (see results for experimental detonation pressure ($P$ exp) and experimental heat of detonation ($Q_{ex}$ exp) in Figure 3). The multi-task multi-fidelity model capable of predicting detonation, sensitivity, and thermodynamic properties of energetic materials simultaneously from basic molecular information and crystal density (MT-NN-all), was found to be the most accurate for several properties. The models presented in this work are designed to facilitate large scale energetic materials screening and are readily applicable in generative based approaches. Future work should explore this type of custom multi-task learning implemented in a graph-based network structure as well as coupling these predictive models with a generative model for ML-driven materials design and exploration.

## Data and Software Availability

The compiled dataset and example code used for model development for this work is available on GitHub at https://github.itap.purdue.edu/StrachanGroup/MultiTaskEM. More complete descriptions of how to develop these types of multi-task neural networks is available on GitHub at https://github.com/Ramprasad-Group/copolymer_informatics.

## Supporting Information

We have provided a detailed summary of the optimized hyperparameters for each model presented in this work.

## Acknowledgments

This research study was sponsored by the Army Research Laboratory and was accomplished under Cooperative Agreement No. W911NF-20-2-0189.